\documentclass[11pt,a4paper]{article}
\usepackage[hyperref]{emnlp-ijcnlp-2019}
\usepackage{times}
\usepackage{latexsym}
\usepackage{graphicx}
\usepackage{subfigure}
\usepackage{url}
\usepackage{float}
\usepackage{amsfonts}       
\usepackage{amsmath}
\usepackage{enumitem}

\aclfinalcopy 

\title{MemeFaceGenerator: Adversarial Synthesis of Chinese Meme-face \\from Natural Sentences}

\author{Yifu Chen$^1\thanks{ \ \ \ Contribution during internship at Tencent.}$, Zongsheng Wang$^2$, Bowen Wu$^2$, Mengyuan Li$^3\footnotemark[1]$, Huan Zhang$^3\footnotemark[1]$, Lin Ma$^4$, \\ \textbf{Feng Liu$^2$, Qihang Feng$^2$, Baoxun Wang$^2$}\\
      $^1$University of Chinese Academy of Sciences, Beijing, China\\
      $^2$Tencent Platform and Content Group\\
      $^3$Peking University, Beijing, China\\
      $^4$Tencent AI Lab\\ 
      {\tt $^1$chenyifu17@mails.ucas.ac.cn}\\
      {\tt $^2$\{jasoawang,jasonbwwu,scarletliu,careyfeng,asulewang\}@tencent.com}\\
      {\tt $^3$\{limengyuan,zhanghuan123\}@pku.edu.cn}\\
      {\tt $^4$forestlma@tencent.com}\\
}

\date{}

\begin{document}
\maketitle
\begin{abstract}
Chinese meme-face is a special kind of internet subculture widely spread in Chinese Social Community Networks. 
It usually consists of a template image modified by some amusing details and a text caption.
In this paper, we present MemeFaceGenerator, a Generative Adversarial Network with the attention module and template information as supplementary signals,
to automatically generate meme-faces from text inputs.
We also develop a web service as system demonstration of meme-face synthesis. 
MemeFaceGenerator has been shown to be capable of generating high-quality meme-faces from random text inputs.
\end{abstract}

\section{Introduction}
\label{section:introduction}

In Chinese Social Community Networks (SNS), 
hand-made meme-faces stand as a new fashion and even a special kind of pop-culture. 
Different from the classic emoji widely applied in the Apps such as WhatsApp\footnote{https://www.whatsapp.com/}, Twitter\footnote{https://twitter.com/}, etc.~\cite{radpour2017conditional,cunha2018shell},
the meme-faces used in the Chinese SNS (e.g., WeChat\footnote{https://weixin.qq.com/}, Weibo\footnote{https://www.weibo.com/}, etc.) 
can be transformed from various meaningful images to express more abstract emotions.

\begin{figure}[!t]
  \centering
  \includegraphics[width=\linewidth]{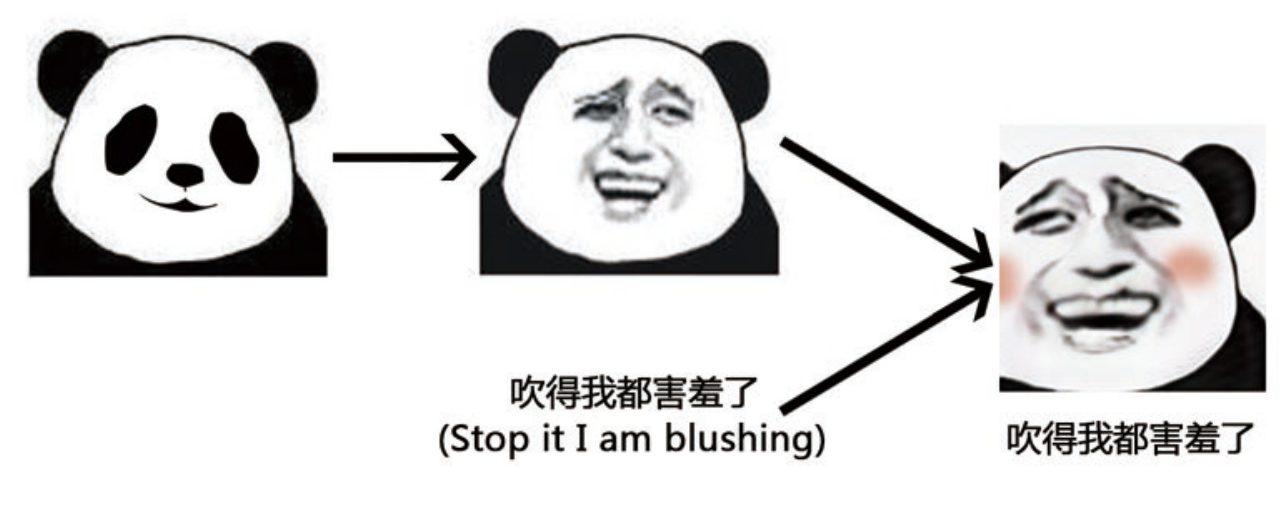}
  \caption{The illustration of the creation of a hand-made meme-face, taking the input \textit{Stop it I am blushing} (translated) as the example.}
  \label{fig:example}
\end{figure}

Manually creating a meme-face is actually a re-creation process on the basis of an existing image from the web,
which is one of the facts that the emotion presenting capacity can be attributed to.
Generally, as illustrated by Figure~\ref{fig:example}, 
the creating procedure of a meme-face consists of the following operations:
a) Picking a meaningful image and changing its details to make it amusing (e.g. switching face to a celebrity);
and b) Based on a given text to be presented, 
further adding or changing some details to align the semantic of the image to the text.
In this entire procedure, computer is only taken as an image editing tool, 
without any inspiration-related function. 

It should be noted that, recently, 
learning to bridge the semantics between image and natural language has become an active research area,
since it has been driving the studies upon multi-modal learning on vision and natural language~\cite{antol2015vqa,reed2016learning,yang2016stacked,reed2016learningdeep}. 
Especially, after the emerging and widely spreading of the Generative Adversarial Networks (GANs)~\cite{goodfellow2014generative,mirza2014conditional},
directly generating images or emojis according to the given natural language descriptions via the End-to-End learning becomes achievable~\cite{reed2016generative,zhang2017stackgan++,zhang2017stackgan,radpour2017conditional,xu2018attngan}.

\begin{figure*}[t!]
	\centering
	\includegraphics[width=\linewidth]{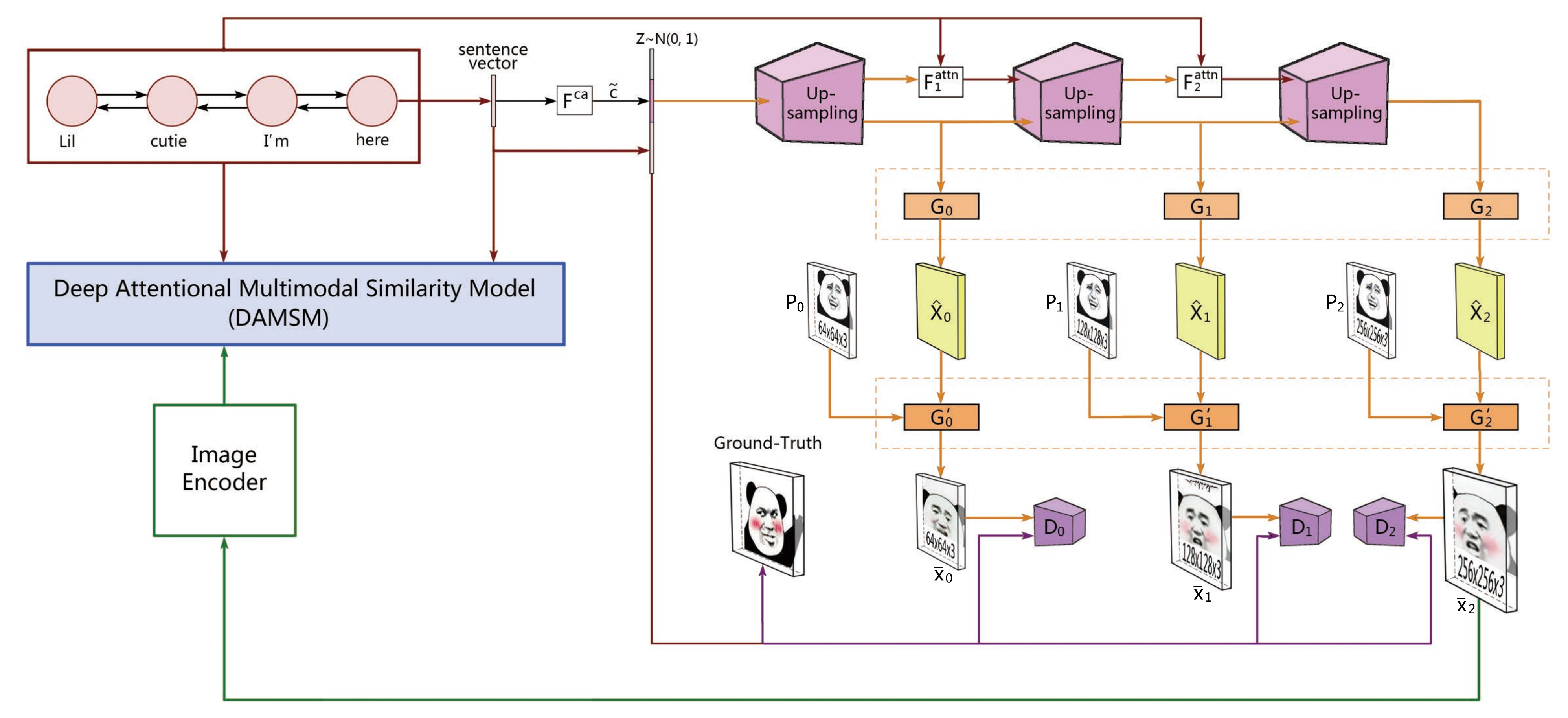}
	\caption{The architecture of the proposed End-to-End Meme-Face Generator. Due to the generation difficulty brought by the implicit semantic relevance of the text and image within a Chinese meme-face, we propose to introduce templates into each generation step.}
	\label{fig:model}
\end{figure*}

This paper aims at generating the meme-face according to the given text input directly,
which is theoretically similar with those tasks discussed in \cite{reed2016generative,zhang2017stackgan++,zhang2017stackgan,dash2017tac,xu2018attngan}, 
with the only difference on the type of data.
However, such difference brings the notable challenge to the model, 
since the image part of a hand-made Chinese meme-face is often only closely related to the text part in terms of semantics.
For the images in the classic datasets such as the widely used COCO dataset~\cite{lin2014microsoft} and the CUB dataset~\cite{wah2011caltech}, by contrast,
the captions provide the descriptions of images by maintaining the lexicon-object matching relationships. 
Overall, the semantics relationship between the text and the image part of such meme-face is beyond the lexicon-object correspondence.

For the Chinese meme-face generation task, 
this paper applies the GAN architecture with the attention module to exploit and capture the complicated relationships of texts and image regions~\cite{xu2018attngan}.
Especially, so as to address the latent semantic relevance to enhance generation, 
this paper proposes to adopt basic patterns in meme-face templates as the supplementary signal, 
which enforces the generator to focus on the modification of the essential local regions to semantically match the given text caption,
with the majority of the pattern fixed. 
The demonstration of our system is released online,
with details presented in the rest sections.

\section{Approach}

\subsection{Model Overview}

Based on the current progress on text to image generation models, we propose a GAN architecture with the attention module named MemeFaceGenerator to generate a meme-face from a pattern representing the semantics of the given text.
As shown in Figure~\ref{fig:model}, stacked attentional generative network~\cite{xu2018attngan} is first employed to model the text into visual-semantic representations $\hat{x_i}$,
where $F^{ca}$ is the Conditioning Augmentation, $F^{attn}_i$ stands for the attention model for obtaining new sentence embedding at the $i^{th}$ stage, and $G_i$ represents the $i^{th}$ generators of the AttnGAN.
Based on such representations and different scaled patterns, the editing component $G'_i$ is proposed to generate the text aligned and pattern closed new meme-faces.
Let $P_i$ stands for the pattern at stage-$i$, network $G'_i$ down-samples $P_i$ and $\hat{x_i}$ into representations with the same dimension, then concatenates them and performs a Multilayer Perceptron (MLP) to integrate the information from both the text and the pattern.
Finally, $G'_i$ utilizes a series of up-sampling blocks to generate images of small-to-large scales ($\bar{x}_0, \bar{x}_1, ..., \bar{x}_m$; where $m$ is the number of stages). 
Beside the generator $G$, the architectures of $m$ discriminators ($D_0, D_1, ..., D_m$) and the text-image matching network Deep Attentional Multimodal Similarity (DAMSM) are inherited from the AttnGAN.

Correspondingly, the objective function of the whole generator $G$ is defined as:
\begin{equation}
\begin{aligned}
\label{equ:loss_G}
    \mathcal{L}&=\mathcal{L}_{G} + \mathcal{L}_{DAMSM} \\
    &=\sum_{i=0}^{m-1} (-\frac{1}{2} \mathbb{E}_{x_{i} \sim p_{G'_{i}}}\left[\log \left(D_{i}\left(\bar{x}_{i}\right)\right]\right. - \\
    &\frac{1}{2} \mathbb{E}_{\bar{x}_{i} \sim p_{G'_{i}}}\left[\log \left(D_{i}\left(\bar{x}_{i}, \hat{c}\right)\right]\right.) + \mathcal{L}_{DAMSM}
\end{aligned}
\end{equation}
where $\hat{c}$ is the sentence vector modeled by a biLSTM layer, and the $\mathcal{L}_{DAMSM}$ loss is computed by a pretrained DAMSM on the Chinese meme-face dataset to evaluate the matching degree between the given text and generated images.
Overall, the following objective function is performed for the training procedure:
\begin{equation}
\begin{aligned}
\label{equ:loss_all}
\min _{G} &\max _{D} V(D, G)= \mathbb{E}_{x \sim p_{\text {data}}}[\log D(x)]+\\ 
& \mathbb{E}_{z}[\log (1-D(G(z, P))) + \mathcal{L}_{DAMSM}]
\end{aligned}
\end{equation}
where $x$ is a read image from the true data distribution $p_{data}$ which belonging to the class represented by the pattern $P$, and $z$ is the noise vector sampled from a uniform distribution.

\subsection{Data Preparation}


As part of experiment preparation, 56,710 meme-faces of various kinds are first collected online, and their text captions are automatically extracted via OCR engine.
To obtain those representative and popular meme-faces, we further implement a series of data cleaning based on the categories of meme-face and the words in text caption.
More specifically:

\begin{itemize}
\item Since most Chinese meme-faces are made based on a small set of templates (such as the panda face in Figure~\ref{fig:example}), clustering is used to find meme-faces from same templates.
Here we use a pre-trained inception-v3 network~\cite{szegedy2016rethinking} to first compute image-vectors from all meme-faces,
and then use k-means for image clustering.
In total 33 representative categories of meme-faces are obtained after removing the outliers. 
\item To refine the quality of text captions in the dataset, 
a language model (trained on text captions) is used to filter those with too little information (e.g. \textit{Yeap}, \textit{Harry Up}, \textit{What}) by setting a proper range of perplexity. 
In addition, we also set a constraint on the length of text captions by only reserving those text of length between 3 and 12.
\end{itemize}

Finally, we crop the meme-face pictures to trim away text captions. 
Implementing the above data preparation process gives us in total of 2955 meaningful meme-faces for training (90\%) and testing (10\%). 

\subsection{Training Details}

For model training, our MemeFaceGenerator is optimized using Adam~\cite{kingma2014adam} with the learning rate of 0.0002 using a batch size of 14. 
The model is trained for 200 epochs in total.
Throughout the training process, we update the generator every five epochs and discriminator every epoch. 
All neural-networks are constructed via Pytorch 0.3.0~\cite{paszke2017automatic}.

\begin{figure*}[t!]
	\centering
	\includegraphics[width=\linewidth, height=9cm]{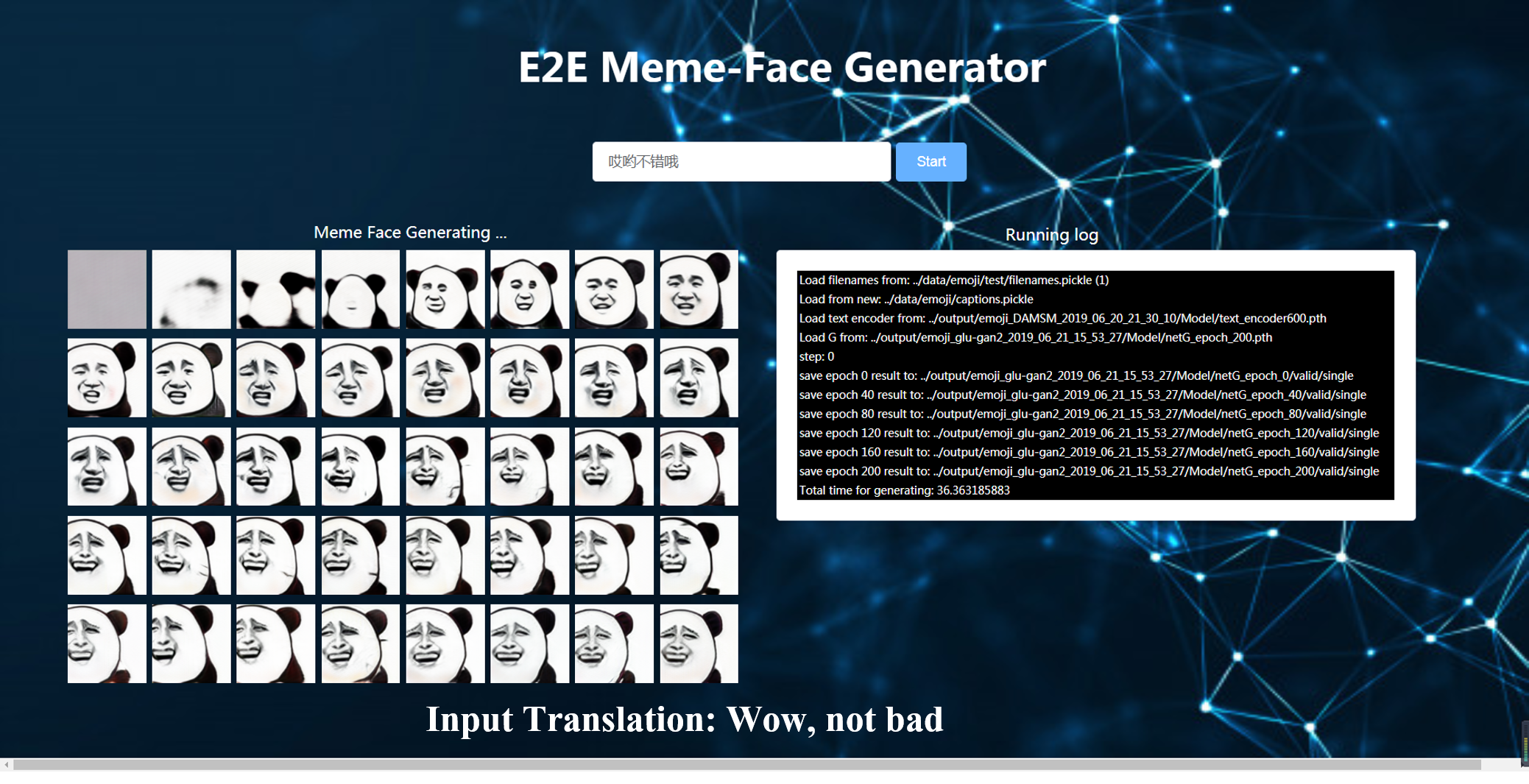}
	\caption{The overview of the system demonstration. Its layout consists of an input box at the top of page, the sequence of generated images for every five epoch one the left side of the page, and corresponding system log on the right hand side of the page}
	\label{fig:system}
\end{figure*}
\section{System Demonstration}

As depicted in Figure~\ref{fig:system}, the web server~\footnote{The screen recording of the web page can be found at \url{https://drive.google.com/file/d/1ewCLGds681LNtRDdwxPkfApWaAmjF03x/view}} of our system demonstration is constructed using vue and tornado. 
After typing the text in the top input box, the web server first transfers the input text from the front-end into \textit{json} format, and post it to back-end as model input.
The model loads the saved parameters every five epochs till the final round and generates corresponding meme-face using the text as input for each loaded epoch. 
Then we post the generated meme-faces sequentially on the web front-end to demonstrate how does an output meme-face vary throughout the training process, together with the system log containing the time elapse and checkpoint information.
All meme-faces are 256x256 pixels and encoded using \textit{base64} schemes. 

The generated image sequence on the left side of the web page gives us a visual hint on the learning process of MemeFaceGenerator throughout the training.
Taking the case in Figure~\ref{fig:system} as an example, after introducing the template information of panda face, the model rapidly learns the profile and general features of meme-face.
Then it starts to focus on generating more detailed facial expression, and adjust the facial expression to match with the input text gradually.
In the end, MemeFaceGenerator synthesizes a smiling face which is semantically relevant to the input text \textit{Wow, not bad}.

\section{Analysis}
\subsection{Case Study}

\begin{figure*}[t!]
\centering 
\subfigure[]{ 
\begin{minipage}{0.33\linewidth}
\centering                                                        
\includegraphics[scale=0.55]{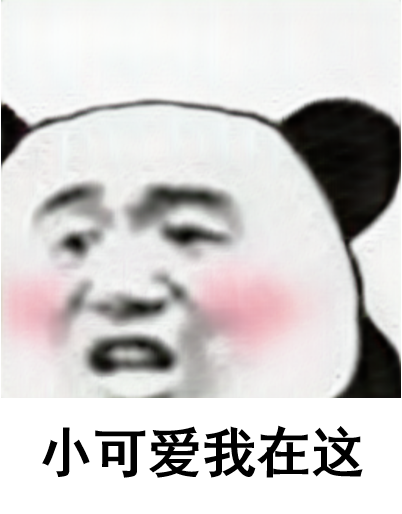}  
\label{fig:case1}
\end{minipage}%
}%
\subfigure[]{ 
\begin{minipage}{0.33\linewidth}
\centering                                                        
\includegraphics[scale=0.55]{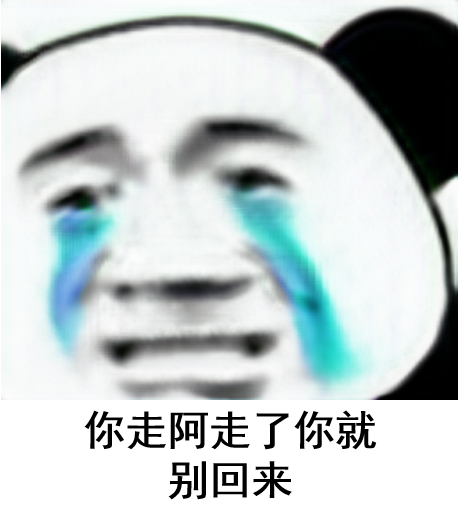}               
\label{fig:case2}
\end{minipage}%
}%
\subfigure[]{ 
\begin{minipage}{0.33\linewidth}
\centering                                                        
\includegraphics[scale=0.55]{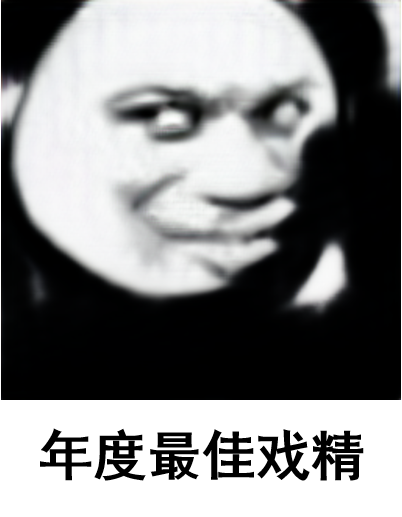}               
\label{fig:case3}
\end{minipage}
}
\subfigure[]{ 
\begin{minipage}{0.33\linewidth}
\centering                                                        
\includegraphics[scale=0.55]{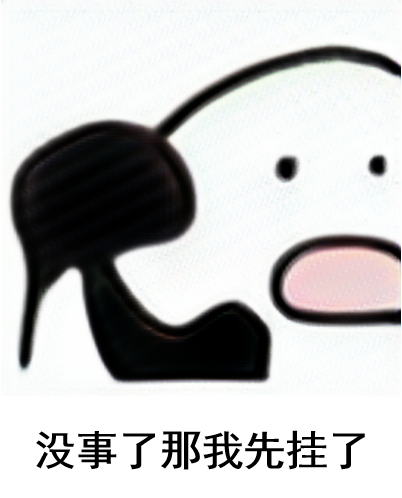}               
\label{fig:case4}
\end{minipage}%
}%
\subfigure[]{ 
\begin{minipage}{0.33\linewidth}
\centering                                                        
\includegraphics[scale=0.55]{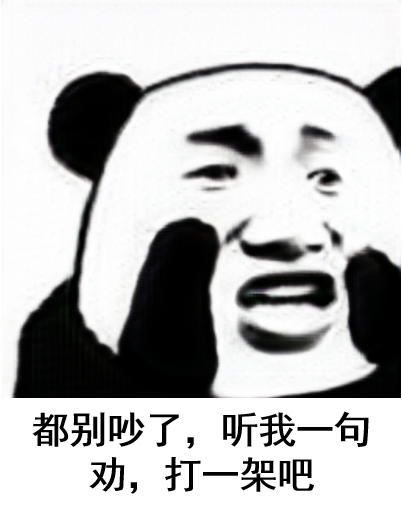}               
\label{fig:case5}
\end{minipage}%
}%
\subfigure[]{ 
\begin{minipage}{0.33\linewidth}
\centering                                                        
\includegraphics[scale=0.5]{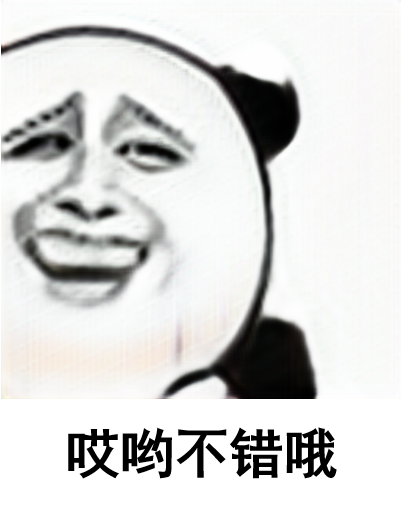}               
\label{fig:case6}
\end{minipage}%
}%
\caption{Six generated meme-faces from test set from text inputs (translated): (a) \textit{lil cutie I'm here.}; (b) \textit{just go, dare you hit the road don't come back.}; (c) \textit{Best drama queen of the year.}; (d) \textit{Nothing important, I'll hang up first.}; (e) \textit{listen to me, stop arguing and put up a fight.}; (f) \textit{Wow, not bad.}}
\label{fig:case full}                                               
\end{figure*}

Ideally, the desired meme-face generator should be capable of generating image consistent with its text caption, 
in other words, the generated image is expected to be semantically-relevant to the text caption.
To verify such capability of our generator, 
in the following section, several generated meme-faces from test set are selected to analyze the acuity and rationality of our proposed MemeFaceGenerator in semantic-relevance capturing.

It can be observed from Figure~\ref{fig:case1} that given the text input \textit{lil cutie I'm here}, the GAN architecture indeed captures the semantics in \textit{lil cutie}, 
reflected by the blushing on cheeks of the image. 
Similarly in Figure~\ref{fig:case4}, our meme-face generator extracts the phrase \textit{hang up (the phone)} from the text input and generates a telephone receiver in the image accordingly.

In addition, rather than capturing relatively straight-forward semantics between the text inputs and images of meme-face, 
Figure~\ref{fig:case2} and Figure~\ref{fig:case5} show signals that the proposed meme-face generator also apprehends more latent semantics between inputs and outputs.
Given the text inputs \textit{just go, dare you hit the road don't come back} and \textit{listen to me, stop arguing and put up a fight},  
the generator extracts the sentiments of sadness and provocation from the text and generates meme-faces with tears and a provoking (shouting) posture to represent the extracted sentiments.

Apart from the face details, the meme-face generator also learns the relevance between faces in different templates and text sentiments through the training.
For example, Figure~\ref{fig:case3} and Figure~\ref{fig:case6} show famous meme-smiling from wrestler D'Angelo Dinero and actor Choi Sung-kook, which are consistent with the ironic and complimentary emotions in their corresponding text inputs \textit{Best drama queen of the year} and \textit{Wow, not bad}.

Overall, the synthesizing results indicate that the visual-semantic representations indeed capture various kinds of semantics from the text input.
With the help of the stacked attentional generative network, MemeFaceGenerator is capable of generating images with details and emotions agreeing with the text inputs.
Such synthesizing results can be directly used in daily communication in social community networks without further manual editing.

\subsection{Numerical Evaluation}

To further analyze the quality of generated results quantitatively,
the generated meme-faces in the test set are cross-evaluated by 3 annotators,
under the following labeling criterion:
\begin{itemize}[label={}]
    \item \textbf{0}: the quality of image is poor or inconsistent with the text caption.
    \item \textbf{1}: the quality of image is acceptable but the image itself is not closely relevant to its text caption. 
    \item \textbf{2}: the image is interesting in terms of its content and matches with its text caption as well. 
\end{itemize}

Overall, the ratios of images labelled as 2, 1 and 0 are 38.8\%, 43.4\% and 17.8\% respectively.
More than 80\% of generated images in the test set are annotated as 1 or higher,
which further verifies the capability of MemeFaceGenerator on modeling the latent semantic relevance between input text and output image.



\section{Conclusions}
In this paper, we exploit the Chinese meme-face generation task using GAN architecture with attention module.
To improve the quality of generated results, meme-face template information is utilized during model training, and precise data-cleaning process is also implemented.
Our proposed MemeFaceGenerator is proved to be able to successfully generate meme-faces consistent with text inputs, it can capture multiple types of semantics between generated images and text captions, 
reflected by the various details and facial expressions in the generated meme-faces.

\bibliography{emnlp-ijcnlp-2019}

\begin{thebibliography}{18}
\expandafter\ifx\csname natexlab\endcsname\relax\def\natexlab#1{#1}\fi

\bibitem[{Antol et~al.(2015)Antol, Agrawal, Lu, Mitchell, Batra,
  Lawrence~Zitnick, and Parikh}]{antol2015vqa}
Stanislaw Antol, Aishwarya Agrawal, Jiasen Lu, Margaret Mitchell, Dhruv Batra,
  C~Lawrence~Zitnick, and Devi Parikh. 2015.
\newblock Vqa: Visual question answering.
\newblock In \emph{Proceedings of the IEEE international conference on computer
  vision}, pages 2425--2433.

\bibitem[{Cunha et~al.(2018)Cunha, Martins, and Machado}]{cunha2018shell}
Jo{\~a}o~Miguel Cunha, Pedro Martins, and Penousal Machado. 2018.
\newblock How shell and horn make a unicorn: Experimenting with visual blending
  in emoji.
\newblock In \emph{ICCC}, pages 145--152.

\bibitem[{Dash et~al.(2017)Dash, Gamboa, Ahmed, Liwicki, and
  Afzal}]{dash2017tac}
Ayushman Dash, John Cristian~Borges Gamboa, Sheraz Ahmed, Marcus Liwicki, and
  Muhammad~Zeshan Afzal. 2017.
\newblock Tac-gan-text conditioned auxiliary classifier generative adversarial
  network.
\newblock \emph{arXiv preprint arXiv:1703.06412}.

\bibitem[{Goodfellow et~al.(2014)Goodfellow, Pouget-Abadie, Mirza, Xu,
  Warde-Farley, Ozair, Courville, and Bengio}]{goodfellow2014generative}
Ian Goodfellow, Jean Pouget-Abadie, Mehdi Mirza, Bing Xu, David Warde-Farley,
  Sherjil Ozair, Aaron Courville, and Yoshua Bengio. 2014.
\newblock Generative adversarial nets.
\newblock In \emph{Advances in neural information processing systems}, pages
  2672--2680.

\bibitem[{Kingma and Ba(2014)}]{kingma2014adam}
Diederik~P Kingma and Jimmy Ba. 2014.
\newblock Adam: A method for stochastic optimization.
\newblock \emph{arXiv preprint arXiv:1412.6980}.

\bibitem[{Lin et~al.(2014)Lin, Maire, Belongie, Hays, Perona, Ramanan,
  Doll{\'a}r, and Zitnick}]{lin2014microsoft}
Tsung-Yi Lin, Michael Maire, Serge Belongie, James Hays, Pietro Perona, Deva
  Ramanan, Piotr Doll{\'a}r, and C~Lawrence Zitnick. 2014.
\newblock Microsoft coco: Common objects in context.
\newblock In \emph{European conference on computer vision}, pages 740--755.
  Springer.

\bibitem[{Mirza and Osindero(2014)}]{mirza2014conditional}
Mehdi Mirza and Simon Osindero. 2014.
\newblock Conditional generative adversarial nets.
\newblock \emph{arXiv preprint arXiv:1411.1784}.

\bibitem[{Paszke et~al.(2017)Paszke, Gross, Chintala, Chanan, Yang, DeVito,
  Lin, Desmaison, Antiga, and Lerer}]{paszke2017automatic}
Adam Paszke, Sam Gross, Soumith Chintala, Gregory Chanan, Edward Yang, Zachary
  DeVito, Zeming Lin, Alban Desmaison, Luca Antiga, and Adam Lerer. 2017.
\newblock Automatic differentiation in pytorch.

\bibitem[{Radpour and Bheda(2017)}]{radpour2017conditional}
Dianna Radpour and Vivek Bheda. 2017.
\newblock Conditional generative adversarial networks for emoji synthesis with
  word embedding manipulation.
\newblock \emph{arXiv preprint arXiv:1712.04421}.

\bibitem[{Reed et~al.(2016{\natexlab{a}})Reed, Akata, Lee, and
  Schiele}]{reed2016learningdeep}
Scott Reed, Zeynep Akata, Honglak Lee, and Bernt Schiele. 2016{\natexlab{a}}.
\newblock Learning deep representations of fine-grained visual descriptions.
\newblock In \emph{Proceedings of the IEEE Conference on Computer Vision and
  Pattern Recognition}, pages 49--58.

\bibitem[{Reed et~al.(2016{\natexlab{b}})Reed, Akata, Yan, Logeswaran, Schiele,
  and Lee}]{reed2016generative}
Scott Reed, Zeynep Akata, Xinchen Yan, Lajanugen Logeswaran, Bernt Schiele, and
  Honglak Lee. 2016{\natexlab{b}}.
\newblock Generative adversarial text to image synthesis.
\newblock \emph{arXiv preprint arXiv:1605.05396}.

\bibitem[{Reed et~al.(2016{\natexlab{c}})Reed, Akata, Mohan, Tenka, Schiele,
  and Lee}]{reed2016learning}
Scott~E Reed, Zeynep Akata, Santosh Mohan, Samuel Tenka, Bernt Schiele, and
  Honglak Lee. 2016{\natexlab{c}}.
\newblock Learning what and where to draw.
\newblock In \emph{Advances in Neural Information Processing Systems}, pages
  217--225.

\bibitem[{Szegedy et~al.(2016)Szegedy, Vanhoucke, Ioffe, Shlens, and
  Wojna}]{szegedy2016rethinking}
Christian Szegedy, Vincent Vanhoucke, Sergey Ioffe, Jon Shlens, and Zbigniew
  Wojna. 2016.
\newblock Rethinking the inception architecture for computer vision.
\newblock In \emph{Proceedings of the IEEE conference on computer vision and
  pattern recognition}, pages 2818--2826.

\bibitem[{Wah et~al.(2011)Wah, Branson, Welinder, Perona, and
  Belongie}]{wah2011caltech}
Catherine Wah, Steve Branson, Peter Welinder, Pietro Perona, and Serge
  Belongie. 2011.
\newblock The caltech-ucsd birds-200-2011 dataset.

\bibitem[{Xu et~al.(2018)Xu, Zhang, Huang, Zhang, Gan, Huang, and
  He}]{xu2018attngan}
Tao Xu, Pengchuan Zhang, Qiuyuan Huang, Han Zhang, Zhe Gan, Xiaolei Huang, and
  Xiaodong He. 2018.
\newblock Attngan: Fine-grained text to image generation with attentional
  generative adversarial networks.
\newblock In \emph{Proceedings of the IEEE Conference on Computer Vision and
  Pattern Recognition}, pages 1316--1324.

\bibitem[{Yang et~al.(2016)Yang, He, Gao, Deng, and Smola}]{yang2016stacked}
Zichao Yang, Xiaodong He, Jianfeng Gao, Li~Deng, and Alex Smola. 2016.
\newblock Stacked attention networks for image question answering.
\newblock In \emph{Proceedings of the IEEE conference on computer vision and
  pattern recognition}, pages 21--29.

\bibitem[{Zhang et~al.(2017{\natexlab{a}})Zhang, Xu, Li, Zhang, Wang, Huang,
  and Metaxas}]{zhang2017stackgan++}
Han Zhang, Tao Xu, Hongsheng Li, Shaoting Zhang, Xiaogang Wang, Xiaolei Huang,
  and Dimitris Metaxas. 2017{\natexlab{a}}.
\newblock Stackgan++: Realistic image synthesis with stacked generative
  adversarial networks.
\newblock \emph{arXiv preprint arXiv:1710.10916}.

\bibitem[{Zhang et~al.(2017{\natexlab{b}})Zhang, Xu, Li, Zhang, Wang, Huang,
  and Metaxas}]{zhang2017stackgan}
Han Zhang, Tao Xu, Hongsheng Li, Shaoting Zhang, Xiaogang Wang, Xiaolei Huang,
  and Dimitris~N Metaxas. 2017{\natexlab{b}}.
\newblock Stackgan: Text to photo-realistic image synthesis with stacked
  generative adversarial networks.
\newblock In \emph{Proceedings of the IEEE International Conference on Computer
  Vision}, pages 5907--5915.

\end{thebibliography}
\bibliographystyle{acl_natbib}

\end{document}